\documentclass[10pt,journal,compsoc]{IEEEtran}
\usepackage{float}
\usepackage{csquotes}
\usepackage{qtree}
\usepackage{tikz}
\usepackage{pdfpages}
\usepackage{array}
\usepackage{amsmath}
\usepackage{algorithm}
\usepackage[noend]{algpseudocode}
\ifCLASSOPTIONcompsoc
  \usepackage[nocompress]{cite}
\else
  \usepackage{cite}
\fi

\ifCLASSINFOpdf
  \usepackage{graphicx}
\else
\fi

\begin{document}
%
\title{Calculating the similarity between words and sentences using a lexical database and corpus statistics}
%
%
%
%

\author{Atish Pawar, Vijay Mago 
\IEEEcompsocitemizethanks{\IEEEcompsocthanksitem A. Pawar and V. Mago are with the Department
of Computer Science, Lakehead University, Thunder Bay,
ON, P7B 5E1.\protect\\
E-mail: \{apawar1,vmago\}@lakeheadu.ca
}
\thanks{}}

%
%

\markboth{IEEE TRANSACTIONS ON KNOWLEDGE AND DATA ENGINEERING}%
{Shell \MakeLowercase{\textit{et al.}}: Calculating the similarity between words and sentences using a lexical database and corpus statistics}
%



\IEEEtitleabstractindextext{%
\begin{abstract}
Calculating the semantic similarity between sentences is a long dealt problem in the area of natural language processing. The semantic analysis field has a crucial role to play in the research related to the text analytics. The semantic similarity differs as the domain of operation differs. In this paper, we present a methodology which deals with this issue by incorporating semantic similarity and corpus statistics. To calculate the semantic similarity between words and sentences, the proposed method follows an edge-based approach using a lexical database. The methodology can be applied in a variety of domains. The methodology has been tested on both benchmark standards and mean human similarity dataset. When tested on these two datasets, it gives highest correlation value for both word and sentence similarity outperforming other similar models. For word similarity, we obtained Pearson correlation coefficient of 0.8753 and for sentence similarity, the correlation obtained is 0.8794. 
\end{abstract}

}

\maketitle
\footnote{This work has been submitted to the IEEE for possible publication. Copyright may be transferred without notice, after which this version may no longer be accessible.}

\IEEEdisplaynontitleabstractindextext

%
\IEEEpeerreviewmaketitle

\IEEEraisesectionheading{\section{Introduction}\label{sec:introduction}}


%
%
%
%
\IEEEPARstart{T}{he} problem of calculating the semantic similarity between two concepts, words or sentences is a long dealt problem in the area of natural language processing. In general, semantic similarity is a measure of conceptual distance between two objects, based on the correspondence of their meanings\cite{lin1998information}. \\
Determination of semantic similarity in natural language processing has a wide range of applications. In internet-related applications, the uses of semantic similarity include estimating relatedness between search engine queries\cite{freitas2011querying} and generating keywords for search engine advertising\cite{abhishek2007keyword}. In biomedical applications, semantic similarity has become a valuable tool for analyzing the results in gene clustering, gene expression and disease gene prioritization\cite{pesquita2009semantic} \cite{lord2003investigating} \cite{pedersen2007measures}. In addition to this, semantic similarity is also beneficial in information retrieval on web \cite{varelas2005semantic}, text summarization\cite{erkan2004lexrank} and text categorization\cite{ko2004improving}. Hence, such applications need to have a robust algorithm to estimate the semantic similarity which can be used across variety of domains.\\
All the applications mentioned above are domain specific and require different algorithms to serve the purpose though the basic idea of calculating the semantic similarity remains the same. To determine the closeness of implications of the objects under comparison, we need some predefined standard measure which readily describes such relatedness of the meanings. The absence of predefined measure makes the problem of comparing definitions, a recursive problem. \\
Lexical databases come into the picture at this point of processing. Lexical databases have connections between words which can be utilized to determine the semantic similarity of the words \cite{fellbaum1998wordnet}. Many approaches have been developed over past few years and proved to be very useful in the area of semantic analysis \cite{baddeley1966short}\cite{resnik1999semantic}\cite{miller1991contextual}\cite{li2006sentence}\cite{lord2003investigating} \cite{jiang1997semantic}. \\
This paper aims to improve existing algorithms and make it robust by integrating it with an corpus of a specific domain. The main contribution of this research is the robust semantic similarity algorithm which outperforms the existing algorithms with respect to the Rubenstein and Goodenough benchmark standard\cite{rubenstein1965contextual}. The application domain of this research is calculating semantic similarity between two \textit{Learning Outcomes} from course description documents. The approach taken to solve this problem is first treating the course objectives as natural language sentences and then introducing domain specific statistics to calculate the simialrity. A separate article will be dedicated to analyze Learning Objectives extracted from different Course Descriptions.\\
The next section reviews some related work. Section 3 elaborates the whole methodology step by step. Section 4 explains the idea of traversal in a lexical database along with an illustrative example in detail. Section 5 contains the result of the algorithm for the 65 noun word pairs from R\&G \cite{rubenstein1965contextual} and the results of the proposed algorithm sentence similarity for the sentence pairs in pilot data set\cite{o2008pilot}. Section 6 discusses the results obtained and compares it with previous methodologies. It also explains the performance of the algorithm. Finally, section 7 presents the outcomes in brief and draws the conclusion. 

\section{Related Work}\label{sec:related_work}
The recent work in the area of natural language processing has contributed valuable solutions to calculate the semantic similarity between words and sentences.  This section reviews some related work to investigate the strengths and limitations of previous methods and to identify the particular difficulties
in computing semantic similarity. Related works can roughly
be classified into following major categories:
\begin{itemize}
\item Word co-occurrence methods
\item Similarity based on a lexical database
\item Method based on web search engine results 
\end{itemize}
Word co-occurrence methods are commonly used in Information
Retrieval (IR) systems \cite{meadow1992text}. This method has word list of
meaningful words and every query is considered as a document. A vector is formed for the query and for documents.
The relevant documents are retrieved based on the similarity
between query vector and document vector [9]. This method
has obvious drawbacks such as:
\begin{itemize}
\item It ignores the word order of the sentence.
\item It does not take into account the meaning of the word in the
context of the sentence.
\end{itemize}
But it has following advantages:
\begin{itemize}
\item It matches documents regardless the size of documents
\item It successfully extracts keywords from documents\cite{matsuo2004keyword}
\end{itemize}
Using the lexical database methodology, the similarity is computed by using
a predefined word hierarchy which has words, meaning, and
relationship with other words which are stored in a tree-like
structure \cite{li2006sentence}. While comparing two words, it takes into account the path distance between the words as well as the depth
of the \textit{subsumer} in the hierarchy. The subsumer refers to the
relative root node concerning the two words in comparison. It also uses a word corpus to calculate the \lq information content\rq of the word which influences the final similarity. This
methodology has the following limitations:
\begin{itemize}
\item The appropriate meaning of the word is not considered
while calculating the similarity, rather it takes the best
matching pair even if the meaning of the word is totally
different in two distinct sentence.
\item The information content of the word form a corpus, differs
from corpus to corpus. Hence, final result differs for every
corpus.
\end{itemize}

The third methodology computes \textit{relatedness} based on web search engine results, utilizes the number of search results \cite{bollegala2007measuring}. This technique doesn't necessarily give the similarity between words as words with opposite meaning frequently occur together on the web pages, hence influencing the final similarity index. We have implemented the methodology to calcuate the Google Similarity Distance \cite{cilibrasi2007google}. The search engines that we used for this study are Google and Bing.  The results obtained from this method are not encouraging for both the search engines.  \\
Overall, above-mentioned methods compute the semantic similarity without considering the context of the word according to the sentence. The proposed algorithm addresses aforementioned issues by disambiguating the words in sentences and forming semantic vectors dynamically for the compared sentences and words. 
\section{The Proposed Methodology}

The proposed methodology considers the text as a sequence of words and deals with all the words in sentences separately according to their semantic and syntactic structure. The information content of the word is related to the frequency of the meaning of the word in a lexical database or a corpus. The method to calculate the semantic similarity between two sentences is divided into four parts: 
\begin{itemize}
  \item Word similarity
  \item Sentence similarity
  \item Word order similarity
\end{itemize}
Fig. 1 depicts the procedure to calculate the similarity between two sentences. Unlike other existing methods that use the fixed structure of vocabulary, the proposed method uses a lexical database to compare the appropriate meaning of the word. A semantic vector is formed for each sentence which contains the weight assigned to each word for every other word from the second sentence in comparison. This step also takes into account the information content of the word, for instance, word frequency from a standard corpus. Semantic similarity is calculated based on two semantic vectors. An order vector is formed for each sentence which considers the syntactic similarity between the sentences. Finally, semantic similarity is calculated based on semantic vectors and order vectors. 
The following section further describes each of the steps in more details. 
\begin{figure}
\label{flowchart}
\centering
  \includegraphics[scale=0.45]{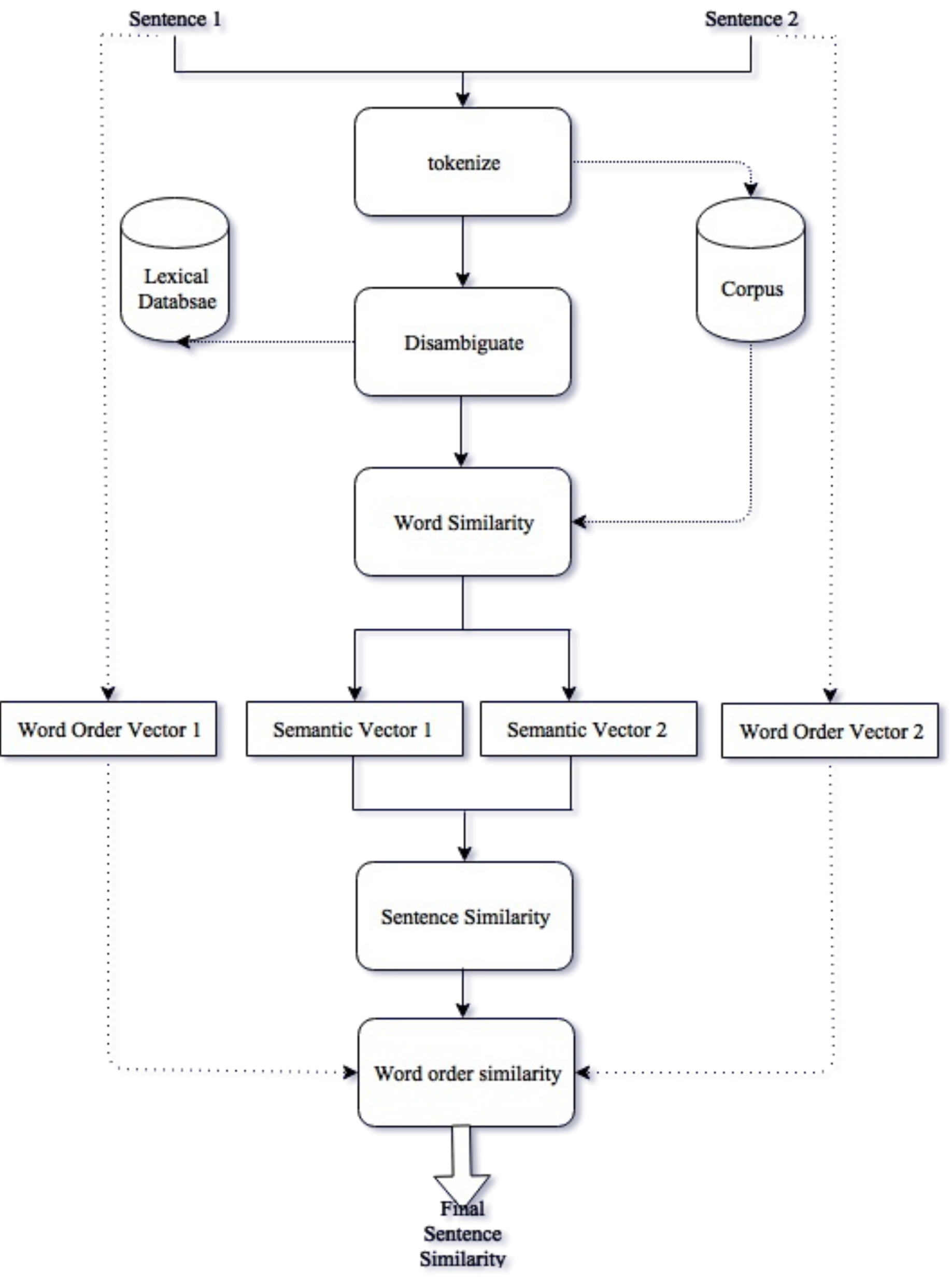}
  \caption{Proposed sentence similarity methodology}
\end{figure} 
\begin{figure*}
\label{bank}
  \includegraphics[width=\textwidth, height=5.5cm]{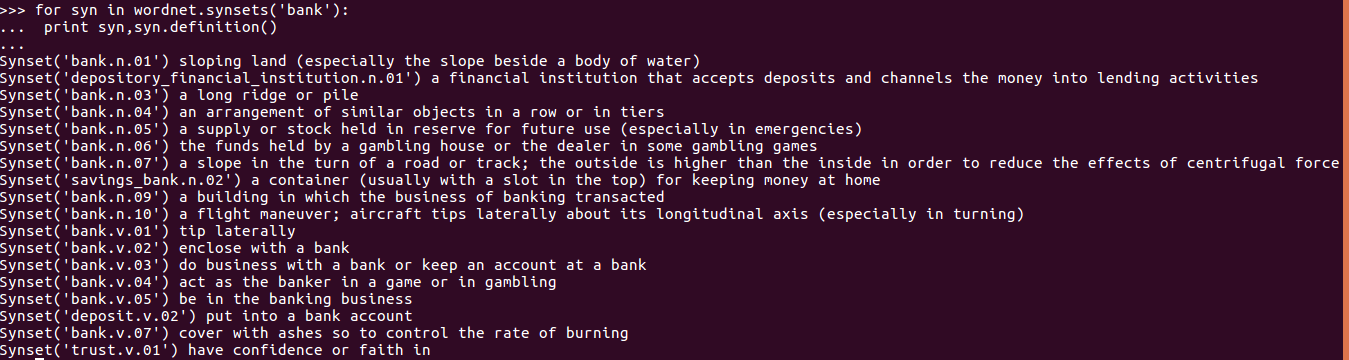}
  \caption{Synsets for the word: bank}
\end{figure*}

\subsection{Word Similarity}
The proposed method uses the sizeable lexical database for the English language, WordNet \cite{miller1995WordNet}, from the Princeton University. Following are the steps involved in computing word similarity:

\subsubsection{Identifying words for comparison}
Before calculating the semantic similarity between words, it
is essential to determine the words for comparison.  We
use word tokenizer and \lq parts of speech tagging technique\rq as
implemented in natural language processing toolkit, NLTK
\cite{bird2006nltk}. This step filters the input sentence and tags the
words into their \lq part of speech\rq(POS) and labels them accordingly. 
As discussed in section \ref{sec:related_work}, WordNet has path relationships between noun-noun and verb-verb only. Such relationships are absent in WordNet for the other parts of speeches. Hence, it is not possible to get a numerical value that represents the link between other parts of speeches except nouns and verbs. Therefore, to reduce the time and space complexity of the algorithm, we only consider nouns and verbs to calculate the similarity. \\
Example: `A voyage is a long journey on a ship or in a spacecraft'\\
Table \ref{table:1} represents the words and corresponding parts of speeches. The parts of speeches are as per the Penn Treebank \cite{marcus1993building}.
\begin{table}
\caption{Parts of speeches}
\label{table:1}
\begin{tabular}{|p{3.5cm}|p{3.5cm}|}
\hline
\textbf{Word} & \textbf{Part of Speech}\\[0.5ex]
\hline
A & DT - Determiner\\[0.5ex]
\hline
voyage & NN - Noun\\[0.5ex]
\hline
is & VBZ - Verb\\[0.5ex]
\hline
a & DT - Determiner\\[0.5ex]
\hline
long & JJ - Adjective\\[0.5ex]
\hline
journey & NN - Noun\\[0.5ex]
\hline
on & IN - Preposition\\[0.5ex]
\hline
a & DT - Determiner\\[0.5ex]
\hline
ship & NN - Noun\\[0.5ex]
\hline
or & CC - Coordinating conjunction\\[0.5ex]
\hline
in & IN - Prepostion\\[0.5ex]
\hline
a & DT - Determiner\\[0.5ex]
\hline
spacecraft & NN - Noun\\[0.5ex]
\hline
\end{tabular}
\end{table}

\subsubsection{Associating word with a sense}
The primary structure of the WordNet is based on \textit{synonymy}. Every word has some synsets according to the meaning of the word in the context of a statement. For example, word: `bank.' Fig. 2 represents all the synsets for the word `bank'.
The distance between \textit{synsets} in comparison varies as we change the meaning of the word. 

Consider an example where we calculate the shortest path distance between words `river' and `bank.' WordNet has only one synset for the word `river'.  We will calculate the path distance between synset of `river' and three synsets of word `bank'.
Table 2 represents the synsets and corresponding definitions for the words `bank' and `river'.

\begin{table}
\caption{Synsets and corresponding definitions from WordNet}
\label{table:2}
\begin{tabular}{|p{3.5cm}|p{3.5cm}|}
\hline
Synset & Definition\\[0.5ex]
\hline
Synset(`river.n.01') & a large natural stream of water (larger than a creek)\\[0.5ex]
\hline
Synset(`bank.n.01') & sloping land (especially the slope beside a body of water)\\[0.5ex]
\hline
Synset(`bank.n.09') & a building in which the business of banking transacted\\[0.5ex]
\hline
Synset(`bank.n.06') & the funds held by a gambling house or the dealer in some gambling games\\[0.5ex]
\hline
\end{tabular}
\end{table}

\begin{table}
\caption{Synsets and corresponding shortest path distances from WordNet}
\label{table:3}
\begin{tabular}{|c|c|}
\hline
Synset Pair & Shortest Path Distance\\[0.5ex]
\hline
Synset(`river.n.01') - Synset(`bank.n.01')& 8\\[0.5ex]
\hline
Synset(`river.n.01') - Synset(`bank.n.09')& 10\\[0.5ex]
\hline
Synset(`river.n.01') - Synset(`bank.n.06')& 11\\[0.5ex]
\hline
\end{tabular}
\end{table}

Shortert distances for synset pairs are represented in Table \ref{table:3}. When comparing two sentences, we have many such word pairs which have multiple synsets. Therefore, not considering the proper synset in context of the sentence, could introduce errors at the early stage of similarity calculation. Hence, sense of the word affects significantly on the overall similarity measure.
Identifying sense of the word is part of the `word sense disambiguation' research area. We use `max similarity'  algorithm, Eq. (1), to perform word sense disambiguation \cite{pedersen2005maximizing} as implemented in Pywsd, an NLTK based Python library \cite{pywsd14}. 
\begin{equation}
{argmax}_{synset(a)}(\sum_{i}^{n}{{max}_{synset(i)}(sim(i,a))}
\end{equation}

\subsubsection{Shortest path distance between synsets}
The following example explains in detail the methodolgy used to calculate the shortest path distance.
\begin{figure}[h]
\label{WordNet}
\includegraphics{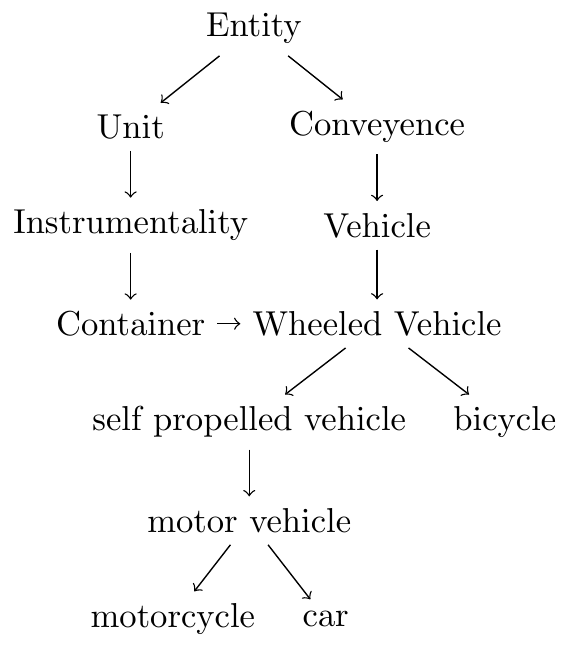}
\caption{Hierarchical structure from WordNet}
\end{figure}

Referring to Fig. 3, consider words: \\
\textit{w1 = motorcycle} and \textit{w2 = car}\\
We are referring to \textit{Synset(`motorcycle.n.01')} for `motorcycle' and \textit{(`car.n.01')} for `car'. \\ The traversal path is :
motorcycle $\,\to\,$ motor vehicle $\,\to\,$ car. Hence, the shortest path distance between motorcycle and car is 2.
In WordNet, the gap between words increases as similarity decreases. We use the previously established monotonically decreasing function \cite{li2006sentence}:\\
\begin{equation}
\label{shortest_path}
f(l)=e^{-\alpha l}
\end{equation}
where \textit{l} is the shortest path distance and $\alpha$ is a constant. The selection of exponential function is to ensure that the value of \textit{f(l)} lies between 0 to 1.

\subsubsection{Hierarchical distribution of words}
\begin{table}
\caption{Synset and corresponding hyponyms from WordNet}
\label{table:4}
\newcolumntype{C}{>{\centering\arraybackslash} m{3.7cm} }
\begin{tabular}{|C|C|}
\hline
Synset & Hyponyms\\[0.5ex]
\hline
Synset(\lq vehicle.n.01\rq) & \shortstack {Synset(`bumper\_car.n.01') \\ Synset(`craft.n.02')\\ Synset(`military\_vehicle.n.01')\\ Synset(`rocket.n.01')\\ Synset(`skibob.n.01')\\ Synset(`sled.n.01')\\ Synset(`steamroller.n.02')\\ Synset(`wheeled\_vehicle.n.01')}\\[0.5ex]
\hline
\end{tabular}
\end{table}   

In WordNet, the primary relationship between the synsets is the super-subordinate relation, also called \textit{hyperonymy}, \textit{hyponymy} or \textit{ISA} relation\cite{miller1995WordNet}. This relationship connects the general concept synsets to the synsets having specific characteristics. 
For example, Table \ref{table:4} represents vehicle and its \textit{hyponyms}.\\
The hyponyms of `vehicle' have more specific properties and represent the particular set, whereas `vehicle' has general properties. Hence, words at the upper layer of the hierarchy have more general features and less semantic information, as compared to words at the lower layer of hierarchy \cite{li2006sentence}. \\
Hierarchical distance plays an important role when the path distances between word pairs are same. For instance,  referring to Fig. 3, consider following word pairs:\\
\textit{car - motorcycle} and \textit{bicycle - self\_propelled\_vehicle}.\\
The shortest path distance between both the pairs is 2, but the pair \textit{car - motorcycle} has more semantic information and specific properties than \textit{bicycle - self\_propelled\_vehicle}. 
Hence, we need to scale up the similarity measure if the word pair \textit{subsume} words at the lower level of the hierarchy and scale down if they subsume words at the upper level of the hierarchy. To include this behavior, we use previously established function\cite{li2006sentence}:\\
\begin{equation}
\label{hierarchical_distance}
g(h)=\dfrac{e^{\beta h}-e^{-\beta h}}{e^{\beta h}+e^{-\beta h}}
\end{equation}
For WordNet, the optimal values of $\alpha$ and $\beta$ are 0.2 and 0.45 respectively as reported previously [8].
\subsection{Information content of the word}
The meaning of the word differs as we change the domain of operation. We can use this behavior of natural language to make the similarity measure domain-specific. Aforementioned is an optional part of the algorithm. It is used to influence the similarity measure if the domain operation is predetermined. 
To illustrate the \textit{Information Content} of the word in action, consider the word: \textit{bank}.
The most frequent meaning of the word \textit{bank} in the context of Potamology (the study of rivers) is \textit{sloping land (especially the slope beside a body of water)}. The most frequent meaning of the word \textit{bank} in the context of Economics would be \textit{a financial institution that accepts deposits and channels the money into lending activities}. \\\\
Used along with the \textit{Word Disambiguation Approach} described in section 3.1.2, the final similarity of the word would be different for every corpus. The corpus belonging to particular domain works as supervised learning data for the algorithm. 
We first disambiguate the whole corpus to get the sense of the word and further calculate the frequency of the particular sense. These statistics for the corpus work as the \textit{knowledge base} for the algorithm. 
Fig. 4 represents the steps involved in the analysis of corpus statistics. 
\begin{figure}[h]
\centering
\label{fig:3}
\includegraphics[scale=0.7]{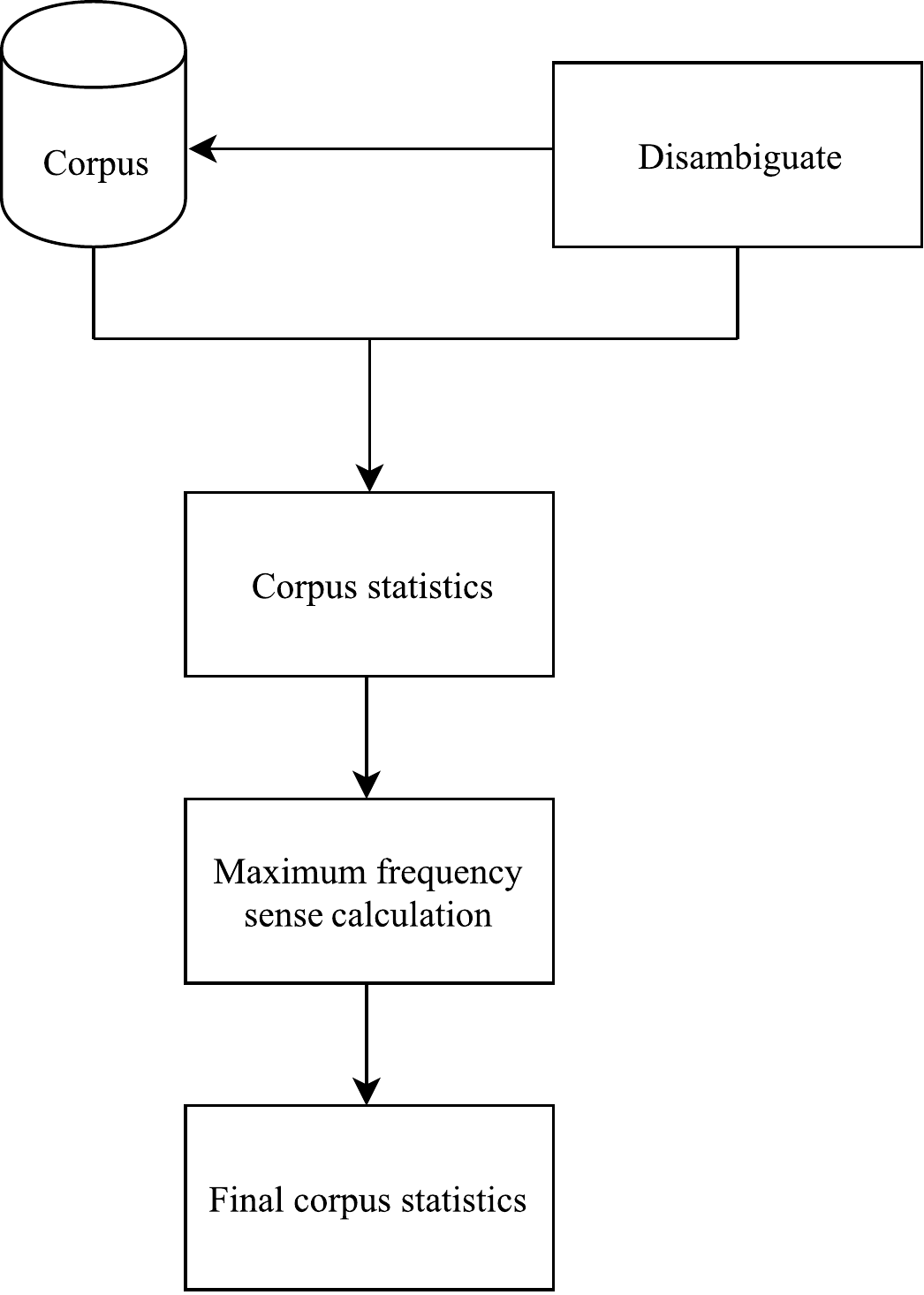}
\caption{Corpus statistics calculation diagram}
\end{figure}
\subsection{Sentences' semantic similarity}
As Li \cite{li2006sentence} states, the meaning of the sentence is reflected by the words in the sentence. Hence, we can use the semantic information from section 3.1 and section 3.2 to calculate the final similarity measure. Previously established methods to estimate the semantic similarity between sentences, use the static approaches like using a precompiled list of words and phrases. The problem with this technique is the precompiled list of words and phrases doesn't necessarily reflect the correct semantic information in the context of compared sentences. \\
The dynamic approach includes the formation of joint word vector which compiles words from sentences and use it as a baseline to form individual vectors. This method introduces inaccuracy for the long sentences and the paragraphs containing multiple sentences. \\
Unlike these methods, our method forms the semantic value vectors for the sentences and aims to keep the size of the semantic value vector minimum. Formation of semantic vector begins after the section 3.1.2. This approach avoids overhead involved to form semantic vectors separately unlike done in previously discussed methods. Also, we eliminate prepositions, conjunctions and interjections in this stage. Hence, these connectives are automatically eliminated from the semantic vector.
We determine the size of the vector, based on the number of tokens from section 3.1.2. Every unit of the semantic vector is initialized to null to void the foundational effect. Initializing semantic vector to a unit positive value discards the negative/null effects, and overall semantic similarity will be a reflection of most similar words in the sentences. Let's see an example. \\ \\
\textit{S1}= ``A jewel is a precious stone used to decorate valuable things that you wear, such as rings or necklaces."\\
\textit{S2}= ``A gem is a jewel or stone that is used in jewellery." \\

List of tagged words for \textit{S1}:\\
\textit{[(`jewel', Synset(`jewel.n.01')), Synset(`jewel.n.02')], }\\
\textit{[(`stone', Synset(`stone.n.02')), Synset(`stone.n.13')], }\\
\textit{[(`used', Synset(`use.v.03')), Synset(`use.v.06')], }\\
\textit{[(`decorate', Synset(`decorate.v.01')), Synset(`dress.v.09')], }\\
\textit{[(`valuable', Synset(`valuable.a.01')), Synset(`valuable.s.02')], }\\
\textit{[(`things', Synset(`thing.n.04')), Synset(`thing.n.12')], }\\
\textit{[(`wear', Synset(`wear.v.01')), Synset(`wear.v.09')], }\\
\textit{[(`rings', Synset(`ring.n.08')), Synset(`band.n.12')], }\\
\textit{[(`necklaces', Synset(`necklace.n.01')), Synset(`necklace.n.01')]}\\ \\
Length of list of tagged words for \textit{S1}: 9
\\ \\
List of tagged words for \textit{S2}:\\
\textit{[(`gem', Synset(`jewel.n.01')), Synset(`jewel.n.01')], }\\
\textit{[(`jewel', Synset(`jewel.n.01')), Synset(`jewel.n.02')], }\\
\textit{[(`stone', Synset(`gem.n.02')), Synset(`stone.n.13')], }\\
\textit{[(`used', Synset(`use.v.03')), Synset(`use.v.06')]}\\ 
\textit{[(`jewellery', Synset(`jewelry.n.01')), Synset(`jewelry.n.01')]}\\ \\
Length of list of tagged words for \textit{S2}: 5\\ \\
We eliminate words like \textit{a, is, to, that, you, such, as, or}; hence further reducing the computing overhead. The formed semantic vectors contain semantic information concerning all the words from both the sentences. 
For example, the semantic vector for \textit{S1} is:\\ 

\textit{V1} = [ 0.99742103,  0.90118787, 0.42189901,  0.0,          0.0,          0.40630945, 0.0,          0.59202,     0.81750916]\\ 

Vector \textit{V1} has semantic information from \textit{S1} as well as from \textit{S2}. Similarly, vector \textit{V2} also has semantic information from \textit{S1} and \textit{S2}. 
To establish a similarity value using two vectors, we use the magnitude of the normalized vectors. 
\begin{equation}
\label{S}
S = ||V1||.||V2||
\end{equation}
We make this method adaptable to longer sentences by introducing a variable($\zeta$) which will be dynamically calculated at runtime. With the utilization of $\zeta$ this method can also be used to compare paragraphs with multiple sentences.
\subsubsection{Determination of $\zeta$}
The words with maximum similarity have more impact on the magnitude of the vector. Using this property, we establish $\zeta$ for the sentences in comparison.  
According to Rubinstein 1965, the benchmark synonymy value of two words is 0.8025 \cite{rubenstein1965contextual}. Using this as a determination standard, we calculate all the cells from \textit{V1} and \textit{V2} with the value greater than 0.8025. $\zeta$ is given by:
\begin{equation}\
\label{zeta}
\zeta = sum(C1, C2)/\gamma
\end{equation}
where $C1$ is count of valid elements in \textit{V1} and $C2$ is count of valid cells in \textit{V2}. $\gamma$ is set to 1.8 to limit the value of similarity in the range of 0 to 1.
Now, using Eq. \ref{S} and Eq. \ref{zeta}, we establish similarity as:
\begin{equation}
\label{S1}
Sim = S/\zeta
\end{equation}
\makeatletter
\def\BState{\State\hskip-\ALG@thistlm}
\makeatother
\begin{algorithm}
\caption{Semantic similarity between sentences}\label{euclid}
\begin{algorithmic}[1]
\Procedure{Sentence\_similarity}{}
\State $\textit{S1 - list of tagged tokens} \gets \textit{disambiguate}$
\State $\textit{S2 - list of tagged tokens} \gets \textit{disambiguate}$
\State $\textit{vector\_length} \gets \textit{max(length(S1),length(S2))}$
\State $\textit{V1},\textit{V2} \gets \textit{vector\_length(null)}$
\State $\textit{V1},\textit{V2} \gets \textit{vector\_length(word\_similarity(S1,S2))}$
\State $\zeta$=0
 \While{$S1\_list\_of\_tagged\_tokens$}
\If{$word\_similarity\_value > \textit{benchmark\_similarity\_value}$} 
\State $C1 \gets \textit{C1+1}$
\EndIf
\EndWhile\label{zetawhile1}

 \While{$\textit{S2}\_list\_of\_tagged\_tokens$}
\If{$word\_similarity\_value>\textit{benchmark\_similarity\_value}$} 
\State $C2 \gets \textit{C2+1}$
\EndIf
\EndWhile\label{zetawhile2}
\State $\zeta \gets {sum(C1, C2)/\gamma}$
\State $S \gets \textit{\begin{math}||V1||.||V2||\end{math}}$
\If {$sum(C1, C2)=0$}
\State $\zeta \gets {vector\_length/2}$
\EndIf
\State $Sim \gets $S$/\zeta$
\EndProcedure
\end{algorithmic}
\end{algorithm}

\subsection{Word Order Similarity}
Along with semantic nature of the sentences, we need to consider the syntactic structure of the sentences too. The word order similarity, simply put, is the aggregation of comparisons of word indices in two sentences. The semantic similarity approach based on words and the lexical database doesn't take into account the grammar of the sentence. Li\cite{li2006sentence} assigns a number to each word in the sentence and forms a word order vector according to their occurrence and similarity. They also consider the semantic similarity value of words to decide the word order vector. If a word from sentence 1 is not present in sentence 2, the number assigned to the index of this word in word order vector corresponds to the word with maximum similarity. This case is not valid always and introduces errors in the final semantic similarity index. 
For the methods which calculate the similarity by chunking the sentence into words, it is not always necessary to decide the word order similarity. For such techniques, the word order similarity actually matters when two sentences contain same words in different order. Otherwise, if the sentences contain different words, the word order similarity should be an optional construct. In the entirely different sentences, word order similarity doesn't impact on the large scale. For such sentences, the impact of word order similarity is negligible as compared to the semantic similarity.
Hence, in our approach, we implement word order similarity as an optional feature.
Consider following classical example:\\
\begin{itemize}
\item \textit{S1}: A quick brown dog jumps over the lazy fox.
\item \textit{S2}: A quick brown fox jumps over the lazy dog.
\end{itemize}

The edge-based approach using lexical database will produce a result showing both \textit{S1} and \textit{S2} are same, but since the words appear in a different order we should scale down the overall similarity as they represent different meaning.
We start with the formation of vectors \textit{V1} and \textit{V2} dynamically for sentences \textit{S1} and \textit{S2} respectively. Initialization of vectors is performed as explained in section 3.3. Instead of forming joint word set, we treat sentences relatively to keep the size of vector minimum. \\
The process starts with the sentence having maximum length. Vector \textit{V1} is formed with respect to sentence 1 and cells in \textit{V1} are initialized to index values of words in \textit{S1} beginning with 1.
Hence \textit{V1} for \textit{S1} is:\\
\begin{math}
\textit{V1} = [1,2,3,4,5,6,7,8,9]
\end{math}\\
Now, we form \textit{V2} concerning \textit{S1} and \textit{S2}. To form \textit{V2},
every word from \textit{S2} is compared with \textit{S1}. If the word from \textit{S2} is absent in \textit{S1}, then the cell in \textit{V2} is filled with the index value of the word in sentence \textit{S2}. If the word from \textit{S2} matches with a word from \textit{S1}, then the index of the word from \textit{S1} is filled in \textit{V2}. \\
In the above example, consider words `fox' and `dog' from sentence 2. The word `fox' from \textit{S2} is present in \textit{S1} at the index 9. Hence, entry for `fox' in \textit{V2} would be 9.
Similarly, the word `dog' form \textit{S2} is present in the \textit{S1} at the index 4. Hence, entry for `dog' in \textit{V2} would be 9. Following the same procedure for all the words, we get \textit{V2} as:\\
\begin{math}
\textit{V2}=[1,2,3,9,5,6,7,8,4]
\end{math}\\
Finally, word order similarity is given by:\\
\begin{equation}
W_s = ||\textit{V1}-\textit{V2}|| / ||\textit{V1}*\textit{V2}||\\
\end{equation}

In this case, $W_s$ is 0.067091.

\section{Implementation using Semantic nets}
The database used to implement the proposed methodology is WordNet and statistical information from WordNet is used calculate the information content of the word. To test the behavior with an external corpus, a small compiled corpus is used. The corpus contained ten sentences belonging to `Chemistry' domain. This section describes the prerequisites to implement the method. 

\subsection{The Database - WordNet}
WordNet is a lexical semantic dictionary available for online and offline use, developed and hosted at Princeton. The version used in this study is WordNet 3.0 which has 117,000 synonymous sets, \textit{Synsets}. Synsets for a word represent the possible meanings of the word when used in a sentence. WordNet currently has synset structure for nouns, verbs, adjectives and adverbs. These lexicons are grouped separately and do not have interconnections; for instance, nouns and verbs are not interlinked. \\
The main relationship connecting the synsets is the super-subordinate(ISA-HASA) relationship. The relation becomes more general as we move up the hierarchy. The root node of all the noun hierarchies is `Entity'. Like nouns, verbs are arranged into hierarchies as well. 
\subsubsection{Shortest path distance and hierarchical distances from WordNet}
The WordNet relations connect the same parts of speeches. Thus, it consists of four subnets of nouns, verbs, adjectives and adverbs respectively. Hence, determining the similarity between cross-domains is not possible. \\
The shortest path distance is calculated by using the tree-like hierarchical structure. To figure the shortest path, we climb up the hierarchy from both the synsets and determine the meeting point which is also a synset. This synset is called \textit{subsumer} of the respective synsets. The shortest path distance equals the hops from one synset to another. \\
We consider the position of subsumer of two synsets to determine the hierarchical distance. Subsumer is found by using the \textit{hyperonymy} (ISA) relation for both the synsets. The algorithm moves up the hierarchy until a common synset is found. This common synset is the subsumer for the synsets in comparison. A set of hypernyms is formed individually for each synset and the intersection of sets contains the subsumer. If the intersection of these sets contain more than one synset, then the synset with the shortest path distance is considered as a subsumer.
\subsubsection{The Information content of the word}
For general purposes, we use the statistical information from WordNet for the information content of the word. WordNet provides the frequency of each synset in the WordNet corpus. This frequency distribution is used in the implementation of section 3.2. 
\subsubsection{Illustrative example}
This section explains in detail the steps involved in the calculation of semantic similarity between two sentences. 
\begin{table}
\caption{L1 compared with L2}
\label{table:5}
\begin{tabular}{|p{3.5cm}|p{3.5cm}|}
\hline
Words & Similarity\\[0.5ex]
\hline
\textbf{gem - jewel} & \textbf{0.908008550956}\\
\hline
gem - stone&0.180732071642\\
\hline
gem - used&0.0\\
\hline
gem - decorate&0.0\\
\hline
gem - valuable&0.0\\
\hline
gem - things&0.284462910289\\
\hline
gem - wear&0.0\\
\hline
gem - rings&0.485032351325\\
\hline
gem - necklaces&0.669319889871\\
\hline
\textbf{jewel - jewel}&\textbf{0.997421032224}\\
\hline
jewel - stone&0.217431543606\\
\hline
jewel - used&0.0\\
\hline
jewel - decorate&0.0\\
\hline
jewel - valuable&0.0\\
\hline
jewel - things&0.406309448212\\
\hline
jewel - wear&0.0\\
\hline
jewel - rings&0.456849659596\\
\hline
jewel - necklaces&0.41718607131\\
\hline
stone - jewel&0.475813717007\\
\hline
\textbf{stone - stone}&\textbf{0.901187866267}\\
\hline
stone - used&0.0\\
\hline
stone - decorate&0.0\\
\hline
stone - valuable&0.0\\
\hline
stone - things&0.198770510639\\
\hline
stone - wear&0.0\\
\hline
stone - rings&0.100270000776\\
\hline
stone - necklaces&0.0856785820827\\
\hline
used - jewel&0.0\\
\hline
used - stone&0.0\\
\hline
\textbf{used - used}&\textbf{0.42189900525}\\
\hline
used - decorate&0.0\\
\hline
used - valuable&0.0\\
\hline
used - things&0.0\\
\hline
used - wear&0.0\\
\hline
used - rings&0.0\\
\hline
used - necklaces&0.0\\
\hline
jewellery - jewel&0.509332774797\\
\hline
jewellery - stone&0.220266070205\\
\hline
jewellery - used&0.0\\
\hline
jewellery - decorate&0.0\\
\hline
jewellery - valuable&0.0\\
\hline
jewellery - things&0.346687374295\\
\hline
jewellery - wear&0.0\\
\hline
jewellery - rings&0.592019999822\\
\hline
\textbf{jewellery - necklaces}&\textbf{0.81750915958}\\
\hline
\end{tabular}
\end{table}

\begin{table}
\caption{L2 compared with L1}
\label{table:6}
\begin{tabular}{|p{3.5cm}|p{3.5cm}|}
\hline
Words & Similarity\\[0.5ex]
\hline
jewel - gem&0.908008550956\\
\hline
\textbf{jewel - jewel}&\textbf{0.997421032224}\\
\hline
jewel - stone&0.475813717007\\
\hline
jewel - used&0.0\\
\hline
jewel - jewellery&0.509332774797\\
\hline
stone - gem&0.180732071642\\
\hline
stone - jewel&0.217431543606\\
\hline
\textbf{stone - stone}&\textbf{0.901187866267}\\
\hline
stone - used&0.0\\
\hline
stone - jewellery&0.220266070205\\
\hline
used - gem&0.0\\
\hline
used - jewel&0.0\\
\hline
used - stone&0.0\\
\hline
\textbf{used - used}&\textbf{0.42189900525}\\
\hline
used - jewellery&0.0\\
\hline
decorate - gem&0.0\\
\hline
decorate - jewel&0.0\\
\hline
decorate - stone&0.0\\
\hline
decorate - used&0.0\\
\hline
decorate - jewellery&0.0\\
\hline
valuable - gem&0.0\\
\hline
valuable - jewel&0.0\\
\hline
valuable - stone&0.0\\
\hline
valuable - used&0.0\\
\hline
valuable - jewellery&0.0\\
\hline
things - gem&0.284462910289\\
\hline
\textbf{things - jewel}&\textbf{0.406309448212}\\
\hline
things - stone&0.198770510639\\
\hline
things - used&0.0\\
\hline
things - jewellery&0.346687374295\\
\hline
wear - gem&0.0\\
\hline
wear - jewel&0.0\\
\hline
wear - stone&0.0\\
\hline
wear - used&0.0\\
\hline
wear - jewellery&0.0\\
\hline
\textbf{rings - gem}&\textbf{0.485032351325}\\
\hline
rings - jewel&0.456849659596\\
\hline
rings - stone&0.100270000776\\
\hline
rings - used&0.0\\
\hline
rings - jewellery&0.592019999822\\
\hline
necklaces - gem&0.669319889871\\
\hline
necklaces - jewel&0.41718607131\\
\hline
necklaces - stone&0.0856785820827\\
\hline
necklaces - used&0.0\\
\hline
\textbf{necklaces - jewellery}&\textbf{0.81750915958}\\
\hline
\end{tabular}
\end{table}   
\begin{itemize}
\item \textit{S1}: A gem is a jewel or stone that is used in jewellery.
\item \textit{S2}: A jewel is a precious stone used to decorate valuable things that you wear, such as rings or necklaces.
\end{itemize}
Following segment contains the parts of speeches and corresponding synsets used to determine the similarity.\\
For \textit{S1} the tagged words are:\\\\
Synset(`jewel.n.01') : a precious or semiprecious stone incorporated into a piece of jewelry\\
Synset(`jewel.n.01') : a precious or semiprecious stone incorporated into a piece of jewelry\\
Synset(`gem.n.02') : a crystalline rock that can be cut and polished for jewelry\\
Synset(`use.v.03') : use up, consume fully\\
Synset(`jewelry.n.01') : an adornment (as a bracelet or ring or necklace) made of precious metals and set with gems (or imitation gems)\\

For \textit{S2} the tagged words are:\\\\
Synset(`jewel.n.01') : a precious or semiprecious stone incorporated into a piece of jewelry\\
Synset(`stone.n.02') : building material consisting of a piece of rock hewn in a definite shape for a special purpose\\
Synset(`use.v.03') : use up, consume fully\\
Synset(`decorate.v.01') : make more attractive by adding ornament, colour, etc.\\
Synset(`valuable.a.01') : having great material or monetary value especially for use or exchange\\
Synset(`thing.n.04') : an artifact\\
Synset(`wear.v.01') : be dressed in\\
Synset(`ring.n.08') : jewelry consisting of a circlet of precious metal (often set with jewels) worn on the finger\\
Synset(`necklace.n.01') : jewelry consisting of a cord or chain (often bearing gems) worn about the neck as an ornament (especially by women)\\ \\
After identifying the synsets for comparison, we find the shortest path distances between all the synsets and take the best matching result to form the semantic vector. 
The intermediate list is formed which contains the words and the identified synsets. L1 and L2 below represent the intermediate lists.\\

 \textit{L1: [(`gem', Synset(`jewel.n.01'))], [(`jewel', Synset(`jewel.n.01'))], [(`stone', Synset(`gem.n.02'))], [(`used', Synset(`use.v.03'))], [(`jewellery', Synset(`jewelry.n.01'))]}\\

 \textit{L2: [(`jewel', Synset(`jewel.n.01'))], [(`stone', Synset(`stone.n.02'))], [(`used', Synset(`use.v.03'))], [(`decorate', Synset(`decorate.v.01'))], [(`valuable', Synset(`valuable.a.01'))], [(`things', Synset(`thing.n.04'))], [(`wear', Synset(`wear.v.01'))], [(`rings', Synset(`ring.n.08'))], [(`necklaces', Synset(`necklace.n.01'))]}
\\\\
Now we begin to form the semantic vectors for \textit{S1} and \textit{S2} by comparing every synset from \textit{L1} with every synset from \textit{L2}. 
The intermediate step here is to determine the size of semantic vector and initialize it to null. In this example, the size of the semantic vector is 9 by referring to the method explained in section 3.3.
The following part contains the cross comparison of \textit{L1} and \textit{L2}.

\begin{table}
\caption{Linear regression parameter values for proposed methodology}
\label{table:7}
\begin{tabular}{|p{3.5cm}|p{3.5cm}|}
\hline
Slope&0.84312603549362108\\
\hline
Intercept&0.017742354112473213\\
\hline
r-value&0.87536955005374539\\
\hline
p-value&1.4816200698817255e-21\\
\hline
stderr&0.058665976202757132\\
\hline
\end{tabular}
\end{table}  
\begin{figure}
\label{vsstd}
  \includegraphics[height=5.3cm]{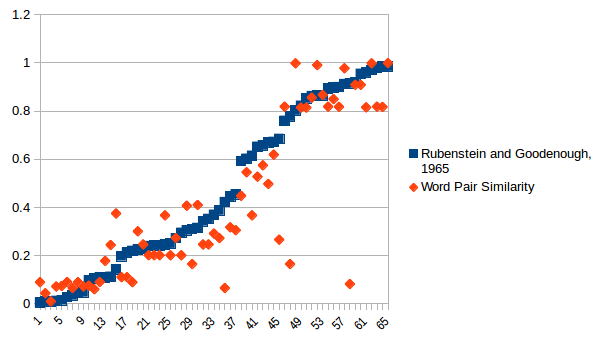}
  \caption{Perfomance of word similarity method vs Standard by Rubenstein and Goodenough}
\end{figure}

\begin{figure}
\label{regression}
  \includegraphics[height=5.3cm]{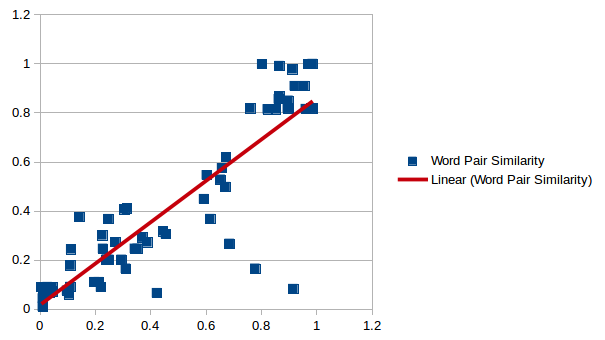}
  \caption{Linear Regression model word similarity method against Standard by Rubenstein and Goodenough}
\end{figure}

\begin{figure}
\label{regression}
  \includegraphics[height=5.3cm]{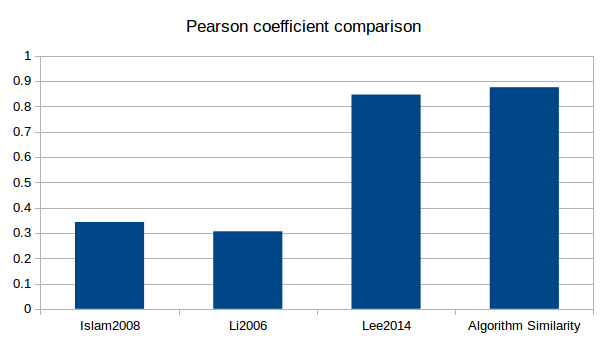}
  \caption{Comparison of linear regressions from various algorithms with R\&G1965}
\end{figure}

\begin{figure}
\label{sentence_regression}
  \includegraphics[height=5.3cm]{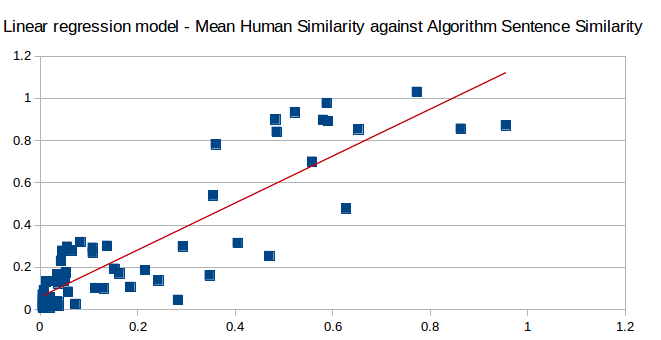}
  \caption{Linear regression model- Mean Human Similarity against Algorithm Sentence Similarity}
\end{figure}

\begin{table*}
\caption{Rubenstein and Goodenough Vs Lee2014 Vs Proposed Algorithm Similarity}
\label{table:7}
\begin{tabular}{|p{3.5cm}|p{3.5cm}|p{3.5cm}|p{3.5cm}|p{3.5cm}|}
\hline
R\&GNo & R\&Gpair & R\&G Similarity & Lee2014 & Proposed Algorithm Similarity	 \\
\hline
1&cord smile &0.005&0.01&0.0899021679\\ 
 \hline 
2&noon string &0.01&0.005&0.0440401486\\ 
 \hline 
3&rooster voyage &0.01&0.0125&0.010051669\\ 
 \hline 
4&fruit furnace &0.0125&0.0475&0.0720444643\\ 
 \hline 
5&autograph shore &0.015&0.005&0.0742552483\\ 
 \hline 
6&automobile wizard &0.0275&0.02&0.0906955651\\ 
 \hline 
7&mound stove &0.035&0.005&0.0656419906\\ 
 \hline 
8&grin implement&0.045 &0.005&0.0899021679\\ 
 \hline 
9&asylum fruit &0.0475&0.005&0.0720444643\\ 
 \hline 
10&asylum monk &0.0975&0.0375&0.0757289762\\ 
 \hline 
11&graveyard madhouse&0.105 &0.0225&0.0607950554\\ 
 \hline 
12&boy rooster &0.11&0.0075&0.0907164485\\ 
 \hline 
13&glass magician &0.11&0.1075&0.1782144411\\ 
 \hline 
14&cushion jewel &0.1125&0.0525&0.2443794293\\ 
 \hline 
15&monk slave&0.1425 &0.045&0.3750880747\\ 
 \hline 
16&asylum cemetery &0.1975&0.0375&0.1106378337\\ 
 \hline 
17&coast forest&0.2125 &0.0475&0.1106378337\\ 
 \hline 
18&grin lad &0.22&0.0125&0.0899021679\\ 
 \hline 
19&shore woodland &0.225&0.0825&0.3011198804\\ 
 \hline 
20&monk oracle &0.2275&0.1125&0.2464473057\\ 
 \hline 
21&boy sage&0.24 &0.0425&0.2017739882\\ 
 \hline 
22&automobile cushion &0.2425&0.02&0.2018466921\\ 
 \hline 
23&mound shore &0.2425&0.035&0.2018466921\\ 
 \hline 
24&lad wizard&0.2475 &0.0325&0.3673305438\\ 
 \hline 
25&forest graveyard &0.25&0.065&0.2015952767\\ 
 \hline 
26&food rooster &0.2725&0.055&0.2732326922\\ 
 \hline 
27&cemetery woodland &0.295&0.0375&0.2015952767\\ 
 \hline 
28&shore voyage &0.305&0.02&0.4075214431\\ 
 \hline 
29&bird woodland &0.31&0.0125&0.1651985693\\ 
 \hline 
30&coast hill &0.315&0.1&0.4103617321\\ 
 \hline 
31&furnace implement &0.3425&0.05&0.2464473057\\ 
 \hline 
32&crane rooster &0.3525&0.02&0.2465928735\\ 
 \hline 
33&hill woodland &0.37&0.145&0.2918421392\\ 
 \hline 
34&car journey &0.3875&0.0725&0.2730713984\\ 
 \hline 
35&cemetery mound &0.4225&0.0575&0.0656419906\\ 
 \hline 
36&glass jewel &0.445&0.1075&0.3176716099\\ 
 \hline 
37&magician oracle &0.455&0.13&0.3057403627\\ 
 \hline 
38&crane implement &0.5925&0.185&0.4486585394\\ 
 \hline 
39&brother lad &0.6025&0.1275&0.5462290271\\ 
 \hline 
40&sage wizard &0.615&0.1525&0.3675115617\\ 
 \hline 
41&oracle sage &0.6525&0.2825&0.5279307332\\ 
 \hline 
42&bird cock &0.6575&0.035&0.5750838807\\ 
 \hline 
43&bird crane&0.67 &0.1625&0.4978503715\\ 
 \hline 
44&food fruit &0.6725&0.2425&0.6196075053\\ 
 \hline 
45&brother monk &0.685&0.045&0.2664571358\\ 
 \hline 
46&asylum madhouse &0.76&0.215&0.8185286992\\ 
 \hline 
47&furnace stove &0.7775&0.3475&0.1651985693\\ 
 \hline 
48&magician wizard &0.8025&0.355&0.9985079423\\ 
 \hline 
49&hill mound &0.8225&0.2925&0.8148010746\\ 
 \hline 
50&cord string &0.8525&0.47&0.8148010746\\ 
 \hline 
51&glass tumbler &0.8625&0.1375&0.8561402541\\ 
 \hline 
52&grin smile &0.865&0.485&0.9910074537\\ 
 \hline 
53&serf slave &0.865&0.4825&0.8673305438\\ 
 \hline 
54&journey voyage &0.895&0.36&0.8185286992\\ 
 \hline 
55&autograph signature &0.8975&0.405&0.8499457067\\ 
 \hline 
56&coast shore &0.9&0.5875&0.8179120223\\ 
 \hline 
57&forest woodland &0.9125&0.6275&0.9780261147\\ 
 \hline 
58&implement tool &0.915&0.59&0.0822919486\\ 
 \hline 
59&cock rooster &0.92&0.8625&0.9093502924\\ 
 \hline 
60&boy lad &0.955&0.58&0.9093502924\\ 
 \hline 
61&cushion pillow &0.96&0.5225&0.8157293861\\ 
 \hline 
62&cemetery graveyard &0.97&0.7725&0.9985079423\\ 
 \hline 
63&automobile car &0.98&0.5575&0.8185286992\\ 
 \hline 
64&gem jewel &0.985&0.955&0.8175091596\\ 
 \hline 
65&midday noon&0.985&0.6525&0.9993931059\\ 
 \hline 
\end{tabular}
\end{table*}  

\begin{table*}
\caption{Proposed Algorithm Similarity Vs Islam2008 Vs Li2006}
\label{table:8}
\begin{tabular}{|p{3.5cm}|p{3.5cm}|p{3.5cm}|p{3.5cm}|p{3.5cm}|}
\hline
R\&G No & R\&G pair& Proposed Algorithm Similarity& A.Islam2008 &Lietal.2006\\ 
 \hline 
1&cord smile &0.0899021679&0.06&0.33\\ 
 \hline 
5&autograph shore &0.0742552483&0.11&0.29\\ 
 \hline 
9&asylum fruit &0.0720444643&0.07&0.21\\ 
 \hline 
12&boy rooster &0.0907164485&0.16&0.53\\ 
 \hline 
17&coast forest &0.1106378337&0.26&0.36\\ 
 \hline 
21&boy sage &0.2017739882&0.16&0.51\\ 
 \hline 
25&forest graveyard &0.2015952767&0.33&0.55\\ 
 \hline 
29&bird woodland &0.1651985693&0.12&0.33\\ 
 \hline 
33&hill woodland &0.2918421392&0.29&0.59\\ 
 \hline 
37&magician oracle &0.3057403627&0.2&0.44\\ 
 \hline 
41&oracle sage &0.5279307332&0.09&0.43\\ 
 \hline 
47&furnace stove &0.1651985693&0.3&0.72\\ 
 \hline 
48&magician wizard &0.9985079423&0.34&0.65\\ 
 \hline 
49&hill mound &0.8148010746&0.15&0.74\\ 
 \hline 
50&cord string &0.8148010746&0.49&0.68\\ 
 \hline 
51&glass tumbler &0.8561402541&0.28&0.65\\ 
 \hline 
52&grin smile &0.9910074537&0.32&0.49\\ 
 \hline 
53&serf slave &0.8673305438&0.44&0.39\\ 
 \hline 
54&journey voyage &0.8185286992&0.41&0.52\\ 
 \hline 
55&autograph signature &0.8499457067&0.19&0.55\\ 
 \hline 
56&coast shore &0.8179120223&0.47&0.76\\ 
 \hline 
57&forest woodland &0.9780261147&0.26&0.7\\ 
 \hline 
58&implement tool &0.0822919486&0.51&0.75\\ 
 \hline 
59&cock rooster &0.9093502924&0.94&1\\ 
 \hline 
60&boy lad &0.9093502924&0.6&0.66\\ 
 \hline 
61&cushion pillow &0.8157293861&0.29&0.66\\ 
 \hline 
62&cemetery graveyard &0.9985079423&0.51&0.73\\ 
 \hline 
63&automobile car &0.8185286992&0.52&0.64\\ 
 \hline 
64&gem jewel &0.8175091596&0.65&0.83\\ 
 \hline 
65&midday noon&0.9993931059&0.93&1\\ 
 \hline 
\end{tabular}
\end{table*} 
\newcolumntype{P}[1]{>{\centering\arraybackslash}p{#1}}
\begin{table*}
\caption{Sentence Similarity from proposed methodology compared with human mean similarity from Li2006}
\label{table:9}
\begin{tabular}{|P{1cm}|P{6.2cm}|P{6.2cm}|P{2cm}|P{2cm}|}
\hline
R\&G number&Sentence 1&Sentence 2&Mean Human Similarity& Proposed Algorithm Sentence Similarity\\
\hline
1&Cord is strong, thick string.&A smile is the expression that you have on your face when you are pleased or amused, or when you are being friendly.& 0.01	&0.0225\\
\hline
2&A rooster is an adult male chicken.&A voyage is a long journey on a ship or in a spacecraft.&0.005&0.2593\\
\hline
3&Noon is 12 o'clock in the middle of the day.&String is thin rope made of twisted threads, used for tying things together or tying up parcels.& 0.0125&0.03455\\
\hline
4&Fruit or a fruit is something which grows on a tree or bush and which contains seeds or a stone covered by a substance that you can eat. & A furnace is a container or enclosed space in which a very hot fire is made, for example to melt metal, burn rubbish or produce steam.& 0.0475&0.1388\\
\hline
5&An autograph is the signature of someone famous which is specially written for a fan to keep.&The shores or shore of a sea, lake, or wide river is the land along the edge of it.& 0.0050&0.0701\\
\hline
6&An automobile is a car.&In legends and fairy stories, a wizard is a man who has magic powers.& 0.0200&0.0088\\
\hline
7&A mound of something is a large rounded pile of it.&A stove is a piece of equipment which provides heat, either for cooking or for heating a room.& 0.0050&0.4968\\
\hline
8&A grin is a broad smile.& An implement is a tool or other pieces of equipment.& 0.0050& 0.0099\\
\hline 
9&An asylum is a psychiatric hospital.& Fruit or a fruit is something which grows on a tree or bush and which contains seeds or a stone covered by a substance that you can eat.&0.0050 & 0.01456\\
\hline 
10&An asylum is a psychiatric hospital.& A monk is a member of a male religious community that is usually separated from the outside world.&0.0375 & 0.0175\\
\hline 
11&A graveyard is an area of land, sometimes near a church, where dead people are buried.& If you describe a place or situation as a madhouse,you mean that it is full of confusion and noise.& 0.0225& 0.1339\\
\hline 
12&Glass is a hard transparent substance that is used to make things such as windows and bottles.& A magician is a person who entertains people by doing magic tricks.& 0.0075& 0.0911\\
\hline 
13&A boy is a child who will grow up to be a man.& A rooster is an adult male chicken.&0.1075 & 0.2921\\
\hline 
14&A cushion is a fabric case filled with soft material, which you put on a seat to make it more comfortable.&A jewel is a precious stone used to decorate valuable things that you wear, such as rings or necklaces. & 0.0525& 0.1745\\
\hline 
15&A monk is a member of a male religious community that is usually separated from the outside world.&A slave is someone who is the property of another person and has to work for that person. &0.0450 & 0.1394\\
\hline 
16&An asylum is a psychiatric hospital.&A cemetery is a place where dead peoples bodies or their ashes are buried. & 0.375&0.03398 \\
\hline 
17&The coast is an area of land that is next to the sea.&A forest is a large area where trees grow close together. &0.0475 &0.3658 \\
\hline 
18&A grin is a broad smile.&A lad is a young man or boy. &0.0125 & 0.0281\\
\hline 
19&The shores or shore of a sea, lake, or wide river is the land along the edge of it.&Woodland is land with a lot of trees. &0.0825 & 0.3192\\
\hline 
20&A monk is a member of a male religious community that is usually separated from the outside world.&In ancient times, an oracle was a priest or priestess who made statements about future events or about the truth. & 0.1125& 0.1011\\
\hline 
21&A boy is a child who will grow up to be a man.&A sage is a person who is regarded as being very wise. & 0.0425& 0.2305\\
\hline 
22&An automobile is a car.&A cushion is a fabric case filled with soft material, which you put on a seat to make it more comfortable. & 0.0200& 0.0330\\
\hline 
23&A mound of something is a large rounded pile of it.&The shores or shore of a sea, lake, or wide river is the land along the edge of it. & 0.0350& 0.0386\\
\hline 
24&A lad is a young man or boy.&In legends and fairy stories, a wizard is a man who has magic powers. & 0.0325& 0.3939\\
\hline 
25&A forest is a large area where trees grow close together.&A graveyard is an area of land, sometimes near a church, where dead people are buried. & 0.0650&0.2787 \\
\hline 
26&Food is what people and animals eat.&A rooster is an adult male chicken. & 0.0550& 0.2972\\
\hline 
27&A cemetery is a place where dead peoples bodies or their ashes are buried.&Woodland is land with a lot of trees. &0.0375 & 0.1240\\
\hline 
28&The shores or shore of a sea, lake, or wide river is the land along the edge of it.&A voyage is a long journey on a ship or in a spacecraft. & 0.0200& 0.0304\\
\hline 
29&A bird is a creature with feathers and wings, females lay eggs, and most birds can fly.&Woodland is land with a lot of trees. & 0.0125& 0.1334\\
\hline 
30&The coast is an area of land that is next to the sea.&A hill is an area of land that is higher than the land that surrounds it. & 0.1000& 0.8032\\
\hline 
31&A furnace is a container or enclosed space in which a very hot fire is made, for example to melt metal, burn rubbish or produce steam.&An implement is a tool or other piece of equipment. & 0.0500& 0.1408\\
\hline 
32&A crane is a large machine that moves heavy things by lifting them in the air.&A rooster is an adult male chicken. & 0.0200&0.0564 \\
\hline 
33&A hill is an area of land that is higher than the land that surrounds it.&Woodland is land with a lot of trees. & 0.1450& 0.7619\\
\hline 

\end{tabular}
\end{table*} 

\newcolumntype{P}[1]{>{\centering\arraybackslash}p{#1}}
\begin{table*}
\caption{Sentence Similarity from proposed methodology compared with human mean similarity from Li2006 (Continued from previous page)}
\label{table:9}
\begin{tabular}{|P{1cm}|P{6.2cm}|P{6.2cm}|P{2cm}|P{2cm}|}
\hline
R\&G Number&Sentence 1&Sentence 2&Mean Human Similarity&Proposed Algorithm Sentence Similarity\\
\hline
34&A car is a motor vehicle with room for a small number of passengers.&When you make a journey, you travel from one place to another. & 0.0725& 0.02610\\
\hline 
35&A cemetery is a place where dead peoples bodies or their ashes are buried.&A mound of something is a large rounded pile of it. & 0.0575& 0.0842\\
\hline 
36&Glass is a hard transparent substance that is used to make things such as windows and bottles.&A jewel is a precious stone used to decorate valuable things that you wear, such as rings or necklaces. & 0.1075& 0.2692\\
\hline 
37&A magician is a person who entertains people by doing magic tricks.&In ancient times, an oracle was a priest or priestess who made statements about future events or about the truth. & 0.1300& 0.1000\\
\hline
38&A crane is a large machine that moves heavy things by lifting them in the air.&An implement is a tool or other piece of equipment. & 0.1850& 0.1060\\
\hline
39&Your brother is a boy or a man who has the same parents as you.&A lad is a young man or boy.
& 0.1275& 0.9615\\
 \hline
40&A sage is a person who is regarded as being very wise.&In legends and fairy stories, a wizard is a man who has magic powers.& 0.1525& 0.1920\\
\hline 
41&In ancient times, an oracle was a priest or priestess who made statements about future events or about the truth.&A sage is a person who is regarded as being very wise.
 & 0.2825& 0.0452\\
\hline 
42&A bird is a creature with feathers and wings, females lay eggs, and most birds can fly.&A crane is a large machine that moves heavy things by lifting them in the air.
& 0.0350& 0.1660\\
\hline 
43&A bird is a creature with feathers and wings, females lay eggs, and most birds can fly.&A cock is an adult male chicken.
 & 0.1625& 0.1704\\
\hline 
44&Food is what people and animals eat.&Fruit or a fruit is something which grows on a tree or bush and which contains seeds or a stone covered by a substance that you can eat. & 0.2425& 0.1379\\
\hline 
45&Your brother is a boy or a man who has the same parents as you.&A monk is a member of a male religious community that is usually separated from the outside world. &0.0450 & 0.2780\\
\hline 
46&An asylum is a psychiatric hospital.&If you describe a place or situation as a madhouse, you mean that it is full of confusion and noise. & 0.2150& 0.1860\\
\hline

47&A furnace is a container or enclosed space in which a very hot fire is made, for example, to melt metal, burn rubbish, or produce steam.&A stove is a piece of equipment which provides heat, either for cooking or for heating a room. & 0.3475& 0.1613\\
\hline

48&A magician is a person who entertains people by doing magic tricks.&In legends and fairy stories, a wizard is a man who has magic powers. &  0.3550& 0.5399\\
\hline
49& A hill is an area of land that is higher than the land that surrounds it.&A mound of something is a large rounded pile of it.& 0.2925& 0.2986\\
\hline
50& Cord is strong, thick string.&String is thin rope made of twisted threads, used for tying things together or tying up parcels. & 0.4700& 0.2530\\
\hline
51& Glass is a hard transparent substance that is used to make things such as windows and bottles.&A tumbler is a drinking glass with straight sides. & 0.1375& 0.3016\\
\hline
52& A grin is a broad smile.&A smile is the expression that you have on your face when you are pleased or amused, or when you are being friendly. & 0.4850& 0.8419\\
\hline
53&In former times, serfs were a class of people who had to work on a particular persons land and could not leave without that persons permission.&A slave is someone who is the property of another person and has to work for that person. & 0.4825& 0.8896\\
\hline
54& When you make a journey, you travel from one place to another.&A voyage is a long journey on a ship or in a spacecraft. &0.3600 &0.7826 \\
\hline
55& An autograph is the signature of someone famous which is specially written for a fan to keep.&Your signature is your name, written in your own characteristic way, often at the end of a document to indicate that you wrote the document or that you agree with what it says. &0.4050 & 0.3146\\

\hline
56&The coast is an area of land that is next to the sea.&The shores or shore of a sea, lake, or wide river is the land along the edge of it. & 0.5875 &0.9773\\
\hline

57&A forest is a large area where trees grow close together.& Woodland is land with a lot of trees.& 0.6275& 0.4770\\
\hline

58& An implement is a tool or other pieces of equipment.&A tool is any instrument or simple piece of equipment that you hold in your hands and use to do a particular kind of work.& 0.5900&0.8919\\
\hline

59& A cock is an adult male chicken.&A rooster is an adult male chicken.& 0.8625&0.8560\\
\hline

60& A boy is a child who will grow up to be a man.&A lad is a young man or boy.&0.5800 &0.8980\\
\hline
61&A cushion is a fabric case filled with soft material, which you put on a seat to make it more comfortable.&A pillow is a rectangular cushion which you rest your head on when you are in bed.&0.5225 & 0.9340\\
\hline
62&A cemetery is a place where dead peoples bodies or their ashes are buried.&A graveyard is an area of land, sometimes near a church, where dead people are buried.& 0.7725& 1.0\\
\hline
63&An automobile is a car.&A car is a motor vehicle with room for a small number of passengers.& 0.5575& 0.7001\\
\hline
64&Midday is 12 oclock in the middle of the day.& Noon is 12 oclock in the middle of the day. &0.9550 & 0.8726\\
\hline
\end{tabular}
\end{table*}

\newcolumntype{P}[1]{>{\centering\arraybackslash}p{#1}}
\begin{table*}
\caption{Sentence Similarity from proposed methodology compared with human mean similarity from Li2006 (Continued from previous page)}
\label{table:10}
\begin{tabular}{|P{1cm}|P{6.2cm}|P{6.2cm}|P{2cm}|P{2cm}|}
\hline
R\&G Number&Sentence 1&Sentence 2&Mean Human Similarity&Proposed Algorithm Sentence Similarity\\
\hline
65 & A gem is a jewel or stone that is used in jewellery.&A jewel is a precious stone used to decorate valuable things that you wear, such as rings or necklaces.& 0.6525& 0.8536\\
\hline
\end{tabular}
\end{table*}
Cross-comparison with all the words from \textit{S1} and \textit{S2} is essential because if a word from statement \textit{S1} best matches with a word from \textit{S2}, does not necessarily mean that it would be true if the case is reversed. This scenario can be observed with the words \textit{jewel} from Table 5 and \textit{things} from Table 6. \textit{things} best matches with \textit{jewel} with index of 0.4063 whereas \textit{jewel} from Table 5 best matches with \textit{jewel} from Table 6.\\
After getting the similarity values for all the word pairs, we need to determine an index entry for the semantic vector. The entry in the semantic vector for a word is the highest similarity value from the comparison with the words from other sentence. 
For instance, for the word \textit{gem}, from Table 5, the corresponding semantic vector entry is 0.90800855 as it is the maximum of all the compared similarity values. \\
Hence, we get \textit{V1} and \textit{V2} as following:\\
\begin{itemize}
\item \textit{V1}= [ 0.90800855,  0.99742103,  0.90118787,  0.42189901,  0.81750916,  0.0,  0.0,  0.0,  0.0]
\item \textit{V2}= [ 0.99742103,  0.90118787,  0.42189901,  0.0,  0.0,  0.40630945,  0.0,  0.59202,  0.81750916]
\end{itemize}
The intermediate step here is to calculate the dot product of the magnitude of normalized vectors: \textit{V1} and \textit{V2} as explained in section 3.3.\\
\begin{math}
\textit{S} = 3.31974454153\\
\end{math}
The following segment explains the determination of $\zeta$ with reference to section 3.3.1. \\
\textit{C1} for \textit{V1} is 4. \textit{C2} for \textit{V2} is 3.
Hence, $\zeta$ is (4+3)/1.8 = 3.89.\\
Now, the final similarity is\\ \textit{Similarity} = \textit{S}/$\zeta$ = 3.31974454153/3.89 = 0.8534.

\section{Experimental Results}
To evaluate the algorithm, we used a standard dataset which has 65 noun pairs originally measure by Rubenstein and Goodenough \cite{rubenstein1965contextual}. The data has been used in many investigations over the years and has been established as a stable source of the semantic similarity measure. The word similarity obtained in this experiment is assisted by the standard sentences in Pilot Short Text Semantic Similarity Benchmark Data Set by James O'Shea \cite{o2008pilot}. 
The aim of this methodology is to achieve results as close as to the benchmark standard by Rubenstein and Goodenough \cite{rubenstein1965contextual}. The definitions of the words are obtained from the Collins Cobuild dictionary. Our algorithm achieved good Pearson correlation coefficient of 0.8753695501 for word similarity which is cosiderably higher than the existing algorithms. \\
Fig. 5 represents the results for 65 pairs against the R\&G benchmark standard. Fig. 6 represents the linear regression against the standard. The linear regression shows that this algortihm outperforms other similar algorithms. Table 7 shows the values of parameters for linear regression. 
\subsection{Sentence similarity}
Tables 10, 11 and 12 contain the mean human sentence similarity values from Pilot Short Text Semantic Similarity Benchmark Data Set by James O'Shea\cite{o2008pilot}. As Li\cite{li2006sentence} explains, when a survey was conducted by 32 participants to establish a measure for semantic similarity, they were asked to mark the sentences, not the words. Hence, word similarity is compared with the R\&G \cite{rubenstein1965contextual} whereas sentence similarity is compared with mean human similarity. Our algorithm's sentence similarity achieved good Pearson correlation coefficient of 0.8794 with mean human similarity outperforming previous methods. Li\cite{li2006sentence} obtained correlation coefficient of 0.816 and Islam \cite{islam2008semantic} obtained correlation coefficient of 0.853. Out of 65 sentence pairs, 5 pairs were eliminated because of their definitions from Collins Cobuild dictionary\cite{sinclair1987looking}. The reasons and results are discussed in next section.

\section{Discussion}
Our algorithm's similarity measure achieved a good Pearson correlation coefficient of 0.8753 with R\&G word pairs \cite{rubenstein1965contextual}. This performance outperforms all the previous methods. 
Table 8 represents the comparison of similarity from proposed method and Lee \cite{lee2014grammar} with the R\&G. Table 9 depicts the comparison of algorithm similarity against Islam \cite{islam2008semantic} and Li \cite{li2006sentence} for the 30 noun pairs and performs better. \\
For sentence similarity, the pairs \textit{17: coast-forest, 24: lad-wizard, 30: coast-hill, 33: hill-woodland and 39: brother-lad} are not considered. The reason for this is, the definition of these word pairs have more than one common or synonymous words. Hence, the overall sentence similarity does not reflect the true sense of these word pairs as they are rated with low similarity in mean human ratings. For example, the definition of `lad' is given as: `A lad is a young man or boy.' and the definition of `wizard' is: `In legends and fairy stories, a wizard is a man who has magic powers.' Both sentences have similar or closely related words such as: `man-man', `boy-man' and `lad-man'. Hence, these pairs affect overall similarity measure more than the actual words compared `lad-wizard'. 

\section{Conclusions}

This paper presented an approach to calculate the semantic similarity between two words, sentences or paragraphs. The algorithm initially disambiguates both the sentences and tags them in their parts of speeches. The disambiguation approach ensures the right meaning of the word for comparison. The similarity between words is calculated based on a previously established edge-based approach. The information content from a corpus can be used to influence the similarity in particular domain. Semantic vectors containing similarities between words are formed for sentences and further used for sentence similarity calculation. Word order vectors are also formed to calculate the impact of the syntactic structure of the sentences.  Since word order affects less on the overall similarity than that of semantic similarity, word order similarity is weighted to a smaller extent. 
The methodology has been tested on previously established data sets which contain standard results as well as mean human results. Our algorithm achieved good Pearson correlation coefficient of 0.8753 for word similarity concerning the bechmark standard and 0.8794 for sentence similarity with respect to mean human similarity. \\
Future work includes extending the domain of algorithm to analyze Learning Objectives from Course Descriptions, incorporating the algorithm with Bloom's taxonomy will also be considered. Analyzing Learning Objectives requires ontologies and relationship between words belonging to the particular field.


%



\ifCLASSOPTIONcompsoc
  \section*{Acknowledgments}
\else
  \section*{Acknowledgment}
\fi
We would like to acknowledge the financial support provided
by ONCAT(Ontario Council on Articulation and Transfer)through Project Number- 2017-17-LU,without their support
this research would have not been possible. We are also grateful to Salimur Choudhury for his insight on different aspects of this
project; datalab.science team for reviewing and proofreading the paper.

\ifCLASSOPTIONcaptionsoff
  \newpage
\fi



\bibliographystyle{IEEEtran}
\bibliography{semantic_similarity}
%



%

\begin{IEEEbiography}
[{\includegraphics[width=1in,height=1.25in,clip,keepaspectratio]{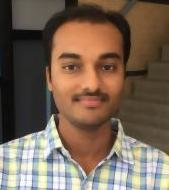}}]{Atish Pawar}
Atish received B.E. degree in computer science and engineering with distinction from Walchand Institute of Technology, India in 2014. He worked for Infosys Technologies from 2014 to 2016. He is currently a graduate student at Lakehead University, Canada. His research interests include machine learning, natural language processing, and artificial intelligence. He is a research assistant at DataScience lab at Lakehead University.
\end{IEEEbiography}



\begin{IEEEbiography}[{\includegraphics[width=1in,height=1.25in,clip,keepaspectratio]{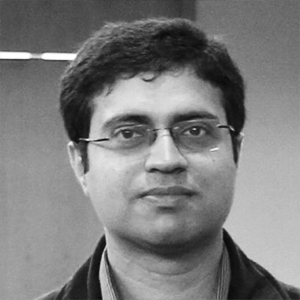}}]{Vijay Mago}
Dr. Vijay. Mago is an Assistant Professor in the Department of Computer Science at Lakehead University in Ontario, where he teaches and conducts research in
areas including decision making in multi-agent environments, probabilistic networks, neural networks, and fuzzy logic-based expert systems. Recently, he has diversified his research to include natural Language Processing, big data and cloud computing. Dr. Mago received his Ph.D. in Computer Science from Panjab University, India in 2010. In 2011 he joined the Modelling of Complex
Social Systems program at the IRMACS Centre of Simon Fraser University before moving on to stints at Fairleigh Dickinson University, University of Memphis and Troy University. He has served on the program committees of many international conferences and workshops. Dr. Mago has published extensively on new methodologies based on soft computing and artificial intelligent techniques to tackle complex systemic problems such as homelessness, obesity, and crime. He currently serves as an associate editor for BMC Medical Informatics and Decision
Making and as co-editor for the Journal of Intelligent Systems.
\end{IEEEbiography}




\end{document}